\documentclass[a4paper,conference]{IEEEtran}
\usepackage{cite}
\ifCLASSINFOpdf
  \usepackage[pdftex]{graphicx}
\else
\fi
\usepackage{amsmath}
\usepackage{algorithmic}
\usepackage{array}
\usepackage{url}
\usepackage{tabularx}
\usepackage[numbers]{natbib}
\usepackage{booktabs}
\usepackage{gensymb}
\usepackage{color}
\usepackage{xcolor}

\newcommand{\shadow}[1]{}

\def\s{\shadow}

\makeatletter
\def\ps@IEEEtitlepagestyle{%
  \def\@oddfoot{\mycopyrightnotice}%
  \def\@evenfoot{}%
}
\def\mycopyrightnotice{%
  {\footnotesize 
  \begin{minipage}
  \textwidth{}
    This work has been submitted to the IEEE for possible publication.\\Copyright may be transferred without notice, after which this version\\ may no longer be accessible.\hfill
  \end{minipage}}% <--- Change here
  \gdef\mycopyrightnotice{}% just in case
}
\begin{document}
\title{\s{ATI-Net:} Attentive Task Interaction Network for Multi-Task Learning}

\author{\IEEEauthorblockN{Dimitrios Sinodinos}
\IEEEauthorblockA{Department of Electrical \& Computer Engineering\\
McGill University\\
Mila - Québec AI Institute\\
Montreal, Canada\\
Email: dimitrios.sinodinos@mail.mcgill.ca}
\and
\IEEEauthorblockN{Narges Armanfard}
\IEEEauthorblockA{Department of Electrical \& Computer Engineering\\
McGill University\\
Mila - Québec AI Institute\\
Montreal, Canada\\
Email: narges.armanfard@mcgill.ca}}


%Telephone: (800) 555--1212, Fax: (888) 555--1212}


\maketitle

\begin{abstract}
Multitask learning (MTL) has recently gained a lot of popularity as a learning
paradigm that can lead to improved per-task performance while also using
fewer per-task model parameters compared to single task learning.
    One of the biggest challenges regarding MTL networks involves how to share features across tasks. To address this challenge, we propose the Attentive Task Interaction Network (ATI-Net). ATI-Net employs knowledge distillation of the latent features for each task, then combines the feature maps to provide improved contextualized information to the decoder. This novel approach to introducing knowledge distillation into an attention based multitask network outperforms state of the art MTL baselines such as the standalone MTAN and PAD-Net, with roughly the same number of model parameters.
\end{abstract}

% \fi
\IEEEpeerreviewmaketitle

\section{Introduction}
% Introduce MTL and provide motivation
The rapid improvements in the performance of machine learning algorithms have led to a surge in their use for a wide range of applications. Particularly for dense image prediction tasks, we see machine learning being used abundantly for applications that are critical to both human health and safety; such as in medical diagnosis, security systems, and autonomous driving. Consequently, improvements in accuracy, inference speed, and parameter efficiency are becoming more imperative for machine learning algorithms.

Multi-Task Learning (MTL) \cite{caruana1997multitask} has demonstrated the ability to enhance performance through improved generalization as a result of leveraging domain-specific knowledge between related tasks \cite{kendall2018multi}. Generally, MTL involves a single network that can learn several tasks by simultaneously optimizing multiple loss functions. Compared to single task networks, MTL networks are potentially more parameter efficient \cite{liu2019end}, leading to faster inference time, and can have improved performance as a result of the inductive bias provided by auxiliary tasks \cite{standley2020tasks}. Recent MTL works for dense prediction tasks \cite{liu2019end,8578175,vandenhende2020mti} can achieve better performance on tasks such as semantic segmentation, depth regression, and surface normals estimation using only one network instead of using several independent  task-specific networks.

% Introduce how attention works with MTL
    %  what improvements can be made to MATN
    %  introduce proposed method
The most commonly addressed challenges with MTL networks include how to share the features, and how to balance the training across tasks. Feature sharing is related to the network architecture. Balancing training typically involves using a loss function with tunable weights assigned to the individual loss functions of each task. The goal is to tune the weights such that one task does not dominate training. Traditionally, a multitask loss function is represented as a weighted sum of the losses for each individual task, as seen in equation \eqref{naive_loss}.

\begin{equation}
\mathcal{L}_{total} = \sum\limits_{i}{\lambda_i\mathcal{L}_i}
\label{naive_loss}
\end{equation}
The majority of MTL works prior to \cite{kendall2018multi} assign weights $\lambda_i$ uniformly for all tasks \cite{eigen2015predicting, 8100062, sermanet2013overfeat}\s{[@Dimitri: CITE SOME RELEVANT PAPERS HERE]}.
Recent works have more sophisticated ways of balancing training across tasks, by tuning the weights based on homoscedastic uncertainty \cite{kendall2018multi, kendall2017uncertainties}, gradient size \cite{chen2018gradnorm}, or based on some performance metrics \cite{guo2018dynamic}. Recent developments in MTL architectures with attention, like the Multi-Task Attention Network (MTAN) \cite{liu2019end}, have shown resilience to different task balancing methods \cite{liu2019end} such as equal weights, uncertainty to weight loss \cite{kendall2018multi}, and dynamic weight averaging (DWA) \cite{liu2019end} inspired by GradNorm \cite{chen2018gradnorm}. Working within an attention based system, the focus of this paper is on architectural modifications to improve performance rather than modifying task balancing schemes.

Architecturally, an MTL network consists of shared layers and task-specific layers. The sharing of the hidden layers are typically done with either hard or soft parameter sharing \cite{9336293, ruder2017overview}. Hard parameter sharing includes sharing the hidden layers between all tasks, and then choosing an arbitrary break-out point that then connects to each of the task-specific heads \cite{9336293}. Soft parameter sharing involves each task having it's own model, where the distance of the model parameters are regularized to keep the parameters across models similar \cite{9336293, ruder2017overview}. MTAN employs a soft parameter sharing scheme that uses attention modules to automatically select the task-specific features from a shared backbone. This will serve as the starting point for developing our proposed architecture.

Another popular topic in MTL involves managing task interactions. Studies \cite{standley2020tasks, 9336293} have shown that choosing which tasks are trained together has a direct impact on the performance of the overall system. MTAN does not have a built-in system for managing task interactions since the set of attention modules for each task only communicate with the shared backbone. 

In this paper, inspired by \cite{8578175, vandenhende2020mti}, we propose the Attentive Task Interaction Network (ATI-Net), which incorporates task-prediction distillation \s{\cite{8578175, vandenhende2020mti}}of the latent features between the shared encoder and decoder of an attention based multi-task network. \s{@Dimitri: The following sentence is not clear: }\s{[Sentence has been removed since it was related to ``version A".]}\s{@Dimitri: More details about the proposed method is needed here. Here you should convince the reviewers that proposed method is novel and helps to address the drawbacks of the related previous methods. Advantages should be included in this paragraph. Basically, this paragraph is about the gist of the paper with a little bit more details compare to the abstract. e.g. See the last two paragraphs of the Intro section in  the MTAN paper. Give an overview of the method, along with its advantages. You want the reader understand the general concept of the method by reading this paragraph.} The incorporation of distillation modules addresses the absence of a task interaction system for the multitask attention network. ATI-Net also takes a different approach to task prediction distillation. Unlike other multitask distillation works \cite{8578175, zhang2018joint, zhang2019pattern, vandenhende2020mti} that will be discussed in the next section, we do not use task prediction distillation as means to refine the final output predictions. Instead, ATI-Net distills low dimension features at an intermediate stage in the network as means to augment the pool of shared features that interact with the task-specific attention modules. Not only does this approach improve performance, but distilling low dimensional features ensures that our architecture remains highly parameter efficient.

% MTAN has no task interaction system
% Padnet is the simplest TPD method and more sophisticated methods use much more parameters, don't all produce results in parallel, and architectures scale poorly with more tasks
% Novelty is how we integrate TDP into a multitask attention network. traditionally, TDP uses the features of the initial prediction, which would be at the attention level in our case. Instead we propose distilling the shared backbone and fusing the output feature maps to provide improved contextual information to the shared decoder. Consequently, the attention masks can select even better features from the shared backbone; which leads to better performance across all tasks. Not only has this shown performance improvements, but the number of parameters increase very slightly. distilling the features at the intermediate stage eliminates the need for introducing larger encoder and decoder heads for each task. This applies to any type of backbone architecture used. Consequently, the model parameters will continue to scale well with more tasks.

In the following sections, there will be an overview of the recent works in attention and task prediction distillation within the context of MTL, followed by a detailed description of the proposed ATI-Net architecture. Afterwards, we will present the experimental conditions related to the dataset, tasks performed, the corresponding evaluation metrics, and the baselines. Finally, there will be a discussion about the corresponding quantitative and qualitative experimental results.

\section{Related Works}
\s{@Dimitri: Note that we don't have to have a related work section. To me, the current related work section can be included in the Introduction section, as its focus is only on MTAN and PAD-NET, the two main pillars of ATI-Net. You may have a more impressive Introduction section if combine sections I and II together. }
% TPD
\subsection{Task-Prediction Distillation}
\begin{figure}
\centering
\includegraphics[width=0.7\linewidth]{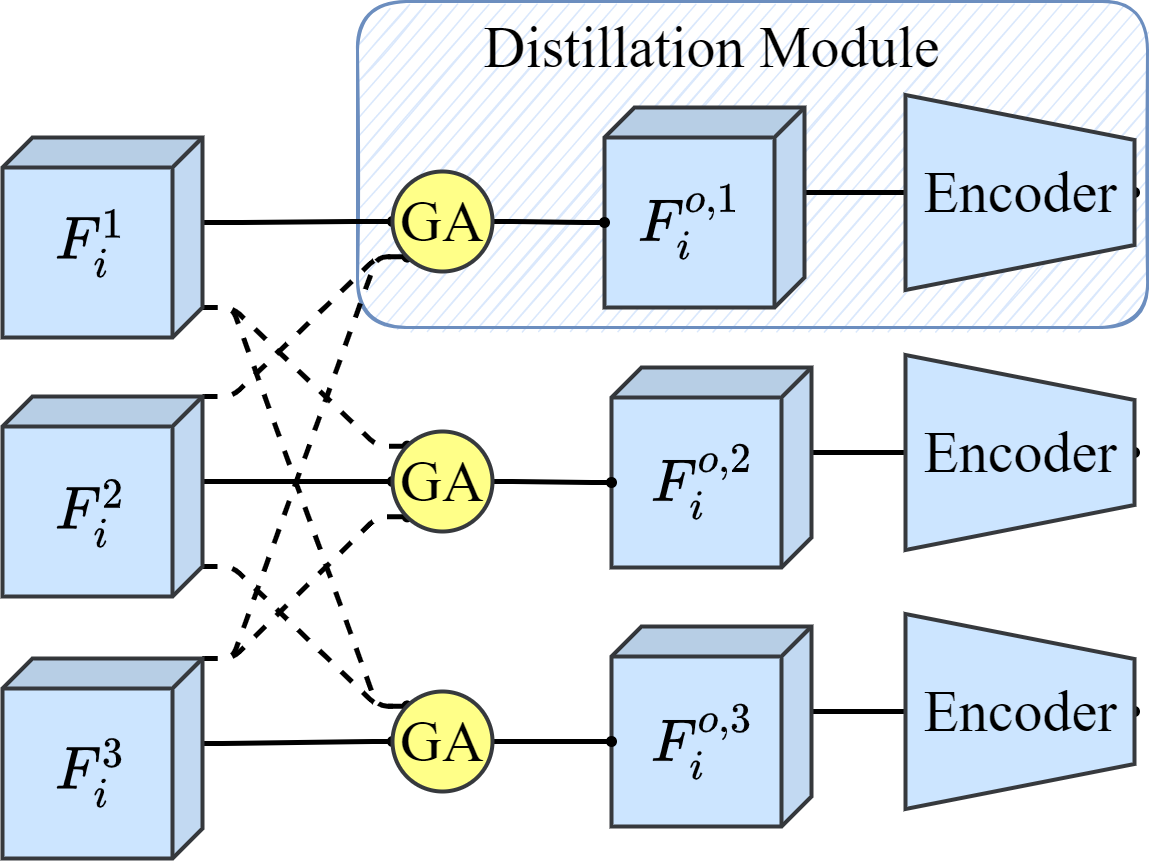}
\caption{Illustration of the distillation module\s{\cite{8578175}}. $F_i^1$, $F_i^2$, $F_i^3$ represent the features of the initial predictions for 3 separate tasks. These features are then propagated through a guided attention mechanism (GA) for passing messages between different predictions. The refined features for each task are denoted as $F_i^{o,1}$, $F_i^{o,2}$, $F_i^{o,3}$. In ATI-Net, we encode these features and pass the resulting average feature map to the shared decoder.}
\label{fig:dist_module}
\end{figure}

Task-prediction distillation involves having the multi-task network make initial predictions for each task, then using the features of these prediction to further refine the output of each task. PAD-Net \cite{8578175} is one of the first state-of-the-art methods to incorporate task prediction distillation. It explores the effect of using three different multi-modal distillation modules, where each use a unique method of distilling the intermediate task outputs to then generate the final task predictions. The best performing module was their ``module C" which uses an attention guided message passing mechanism for information fusion\s{\cite{8578175}}. An illustration of this distillation module and how it's been adapted in our architecture can be seen in Fig. \ref{fig:dist_module}. The benefits of using attention is that it can act as gate function, where the network can automatically learn to focus or to ignore information from features from each task prediction. An important detail about PAD-Net is that the front-end network is first initialized with parameters as a result of pre-training with the ImageNet dataset, and the rest of the network parameters are randomly initialized \cite{8578175}. In addition to initializing with pre-trained parameters, the front-end network is first optimized for the scene-parsing task before the entire multi-task learning begins. Other works that use distillation for MTL include JTRL \cite{zhang2018joint}, PAP-Net \cite{zhang2019pattern}, and MTI-Net \cite{vandenhende2020mti}. JTRL recursively refines task results by using learning experiences from the task interactions that are encapsulated by task attention modules. Learning experiences for each task are then propagated from one task to refine the prediction of another task at each iteration \cite{zhang2018joint,9336293}. PAP-Net also uses a recursive procedure with both cross-task propagation and task-specific propagation of the patterns from the initial predictions \cite{zhang2019pattern}. Contrary to PAD-Net, the patterns involve pixel affinities rather than the actual image features. MTI-Net makes initial task predictions, then uses the features of these predictions at several scales to further refine the output of each task by means of spatial attention in a one-off recursive manner \cite{vandenhende2020mti}.

\begin{figure*}[htb!]
\centering
\includegraphics[width=0.9\linewidth]{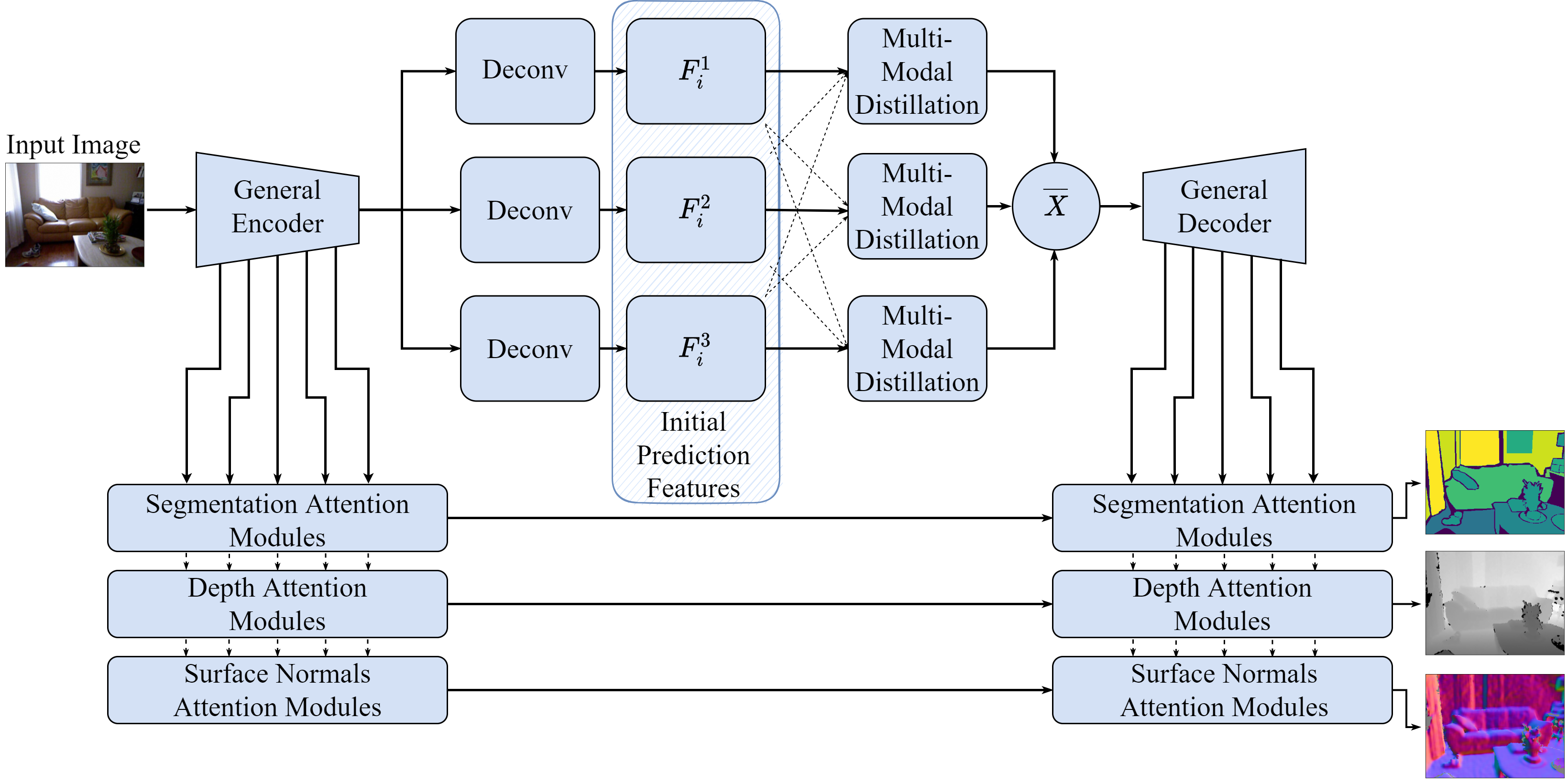}
\caption{ATI-Net architecture which includes task prediction distillation between the general encoder and decoder of a multi task attention network. The latent features from the general encoder are passed through task-specific decoders to generate intermediate prediction features. These features then undergo multi-modal distillation to generate output feature maps for each task. The feature maps are averaged and then passed to the general decoder. Encoder and decoder based attention modules then extract features from the shared backbone to generate the final predictions.}
\label{fig:Task_Dist_B}
\end{figure*}

% MTAN
\subsection{Attention-based Methods}
There exist few methods to employ attention in a multitask learning  framework \cite{9336293}. MTAN \cite{liu2019end} is an attention based multi-task network that consists of a shared backbone encoder-decoder structure that is connected to rows of task-specific attention modules. Each task in the system has it's own dedicated row of attention modules, where the function of the modules is to automatically determine the importance of the shared features from the shared backbone. A benefit of MTAN compared to other soft parameter sharing schemes, like  cross-stitch networks \cite{misra2016cross} or NDDR-CNNs \cite{gao2019nddr}, is that the attention modules are small compared to the general encoder and decoder; making it scale better with more tasks. As previously mentioned, MTAN is also resilient to different task balancing methods. Another important feature is that the network was able to achieve state of the art performance training end-to-end without any pretraining.
\s{@Dimitri: Cite more attention based multi task parers, in this subsection. Related works is for quickly (usually with no detail) introducing a bunch of papers that share similar concepts.} \s{I actually have not been able to find other works that use attention in a multi task setting. Refer to \cite{9336293} which is a recent survey of MTL papers; it also only includes MTAN for multitask attention networks.}

\section{Proposed Method}
One limitation to the MTAN model is that it can only use limited local information to produce the attention mask \cite{9336293}. Our approach improves performance by injecting a task interaction system using task prediction distillation in the latent space.

\subsection{Model Architectures}
\s{@Dimitri: To help the reviewers realizing what parts are proposed by the authors, use the wording like " We propose to ....", instead of just writing everything together not being clear what the current paper's novelties are.}

The aforementioned works on task prediction distillation use various distillation strategies as a means to produce the final output predictions. Consequently, these works are motivated to employ expensive recursive strategies or manipulate the features at multiple scales in order to obtain the best final predictions. Rather than distilling after the prediction layer of an existing network \cite{zhang2018joint,zhang2019pattern,8578175,vandenhende2020mti}, we propose to distill the latent features between the shared encoder and decoder of a multitask attention network. Consequently, ATI-Net provides more refined and contextualized features to the shared decoder. This allows the attention modules to select better features for each task; leading to improved per-task performance. Additionally, since we are not performing task prediction distillation at the end of the network, we are not obliged to operate with features maps corresponding to the full size output resolution. Consequently, we propose generating the initial predictions by passing the low resolution latent features through a series of lightweight dedicated decoder heads to reduce the dimensionality of the feature channels prior to distillation. This drastically reduces the number of parameters introduced to the model. An overview of the proposed ATI-Net for semantic segmentation, depth regression, and surface normals estimation can be seen in Fig. \ref{fig:Task_Dist_B}. The dedicated decoder heads are illustrated by the Deconv blocks. Specifically, our proposed dedicated decoder heads for each task are comprised of a series of 3 deconvolution layers to reduce the channel dimension from 512 to 64. Each deconvolutional layer consists of a 2D convolution with a 1x1 kernel, a batch normalization, and a ReLu activation. The output of each deconvolution operation are the initial prediction features for its corresponding task.

Inspired by \cite{8578175}, the initial prediction features are then refined through Multi-Modal Distillation modules. This involves passing each of the initial prediction features of the auxiliary tasks as inputs to self attention blocks. We define auxiliary tasks for a given task ``t" as all tasks used in the training of the multi-task model except for ``t". In other words, an auxiliary task is a task whose prediction features are used to help further refine the prediction features of another task. For example, the auxiliary tasks for semantic segmentation in our experiments are depth regression and surface normal estimation. The self attention blocks take the predictions of the auxiliary tasks and pass each one through a convolutional layer and a sigmoid activation. The output of the self attention blocks for the auxiliary tasks are then concatenated and summed along the channel dimension, and finally added to the initial prediction features of task ``t". The aforementioned procedure occurs for each of the tasks, and the corresponding refined outputs then needed to be encoded to fit as input to the general decoder. To encode the distilled features, the reverse operation from the decoder heads is performed, which consists of a series of 3 convolutional blocks that increased the channel dimensions from 64 back to 512. This operation is represented by the task-specific encoder blocks in Fig. \ref{fig:dist_module}. The re-encoded feature maps for each task are averaged together, and then fed into the general decoder. This averaging of the feature maps is denoted by the $\bar{X}$ block in Fig. \ref{fig:Task_Dist_B}.

Even with the additional intermediate task-specific decoder-encoder heads, the increase in the number of parameters compared to the original multitask attention network was only 3.3\%. With the enhanced backbone in place, the encoder and decoder based attention modules have an improved pool to select task shared features from. The five arrows connecting the general encoder and decoder to the task-specific attention modules represents the connections to the five attention modules found in each attention block in Fig. \ref{fig:Task_Dist_B}. These connections propagate task shared features, which is represented by the arrows that cascade down to the other sets of task-specific attention modules. The output of the decoder based task-specific attention modules are then passed through  prediction layers to generate the final outputs.

\subsection{Training Configurations}
The distillation modules will only be activated halfway through the entire training. This means that for the first half of the training, the features from the shared encoder will be sent directly to the shared decoder. For the second half of the training, the features from the shared encoder will pass through the Deconv blocks to produce the initial prediction features as seen in Fig. \ref{fig:Task_Dist_B}. \s{@ Dimitri: more clarify the previous sentence. } This is to ensure the modules are actually distilling meaningful features across tasks. At this halfway point, the task-specific attention modules will also be reset to a randomly initialized state. \s{@Dimitri: clarify again here. reset to what? } This allows the attention modules to relearn the meaningful mappings of task shared features from the enhanced backbone.

\section{Experiments}
\subsection{Dataset}
We perform our experiments on the most commonly used benchmark dataset NYUv2 \cite{Silberman:ECCV12}. NYUv2 contains 1449 densely labeled RGB-depth images generated by recordings of indoor scenes using a Microsoft Kinect device. Specifically, following \cite{liu2019end}, we use the raw dataset that includes incomplete depth values for certain images. The train set and validation set are comprised of 795 and 654 images respectively to learn depth regression, surface normal estimation, and 13-class semantic segmentation. The pseudo ground truth surface normals data is from \cite{eigen2015predicting}, which also includes some incomplete values for the same associated pixels from the depth maps. The resolution of the images used are [288$\times$384].

\subsection{Overview of tasks}
\textbf{Semantic Segmentation} involves assigning a class label to every pixel in an image. The training objective is to minimize the depth-wise cross entropy loss between the predicted labels ($\hat{y}$) with the ground truth ($y$) for all $N_S$ pixels. The loss function is formulated as seen in equation \eqref{sem_loss}.

\begin{equation}
    \mathcal{L}_{Semantic} = -\frac{1}{N_S}\sum_{n \epsilon N_S} y_n \text{log}(\hat{y}_n)
\label{sem_loss}
\end{equation}

\textbf{Depth Regression} involves assigning a continuous depth value at each pixel. The training objective is to minimize the $L_1$ norm (i.e. absolute error) between the predicted depth values ($\hat{d}$) and the ground truth ($d$) for all $N_D$ pixels.

\begin{equation}
    \mathcal{L}_{Depth} = \sum_{n \epsilon N_D}||d_n-\hat{d}_n||
\label{abs_err}
\end{equation}

\textbf{Surface Normal Estimation} deals with predicting the surface orientation of the objects present inside a scene. The training objective is to minimize the element-wise dot product between the normalized predicted estimates ($\hat{y}$) with the ground truth ($y$), for all $N_N$ pixels.

\begin{equation}
    \mathcal{L}_{Normals} = -\frac{1}{N_N}\sum_{n \epsilon N_N}y_n\cdot\hat{y}_n
\label{abs_err}
\end{equation}

\subsection{Evaluation Metrics}
As mentioned, the three tasks that will be performed on the NYUv2 dataset are depth regression, surface normal estimation and 13-label semantic segmentation. For depth regression, we will be evaluating the absolute error and the relative error of the estimations with respect to the ground truth values obtained from the Kinect depth sensor. The absolute error is equivalent to the depth loss $\mathcal{L}_{Depth}$, and relative depth error is computed in accordance with equation \eqref{rel_err}, where $d_n$ is the ground truth depth value for pixel $n$, and $\hat{d}_n$ is the depth estimate from our model.

\begin{equation}
    Error_{rel} = \sum_{n \epsilon N_D}\frac{||d_n-\hat{d}_n||}{d_n}
\label{rel_err}
\end{equation}

For surface normal estimation, we will evaluate the mean and median angle distance error with respect to the pseudo ground truths values obtained from \cite{eigen2015predicting}. Specifically, we compute the mean and median of the angular distance between the 3-dimensional predicted and ground truth normal vectors. Angular distance is simply the arccosine of the sum of the element-wise product of the normalized predicted and ground truth vectors, as seen in equation \eqref{angle_distance}, where $\hat{y}_n$ and $y_n$ are the predicted and ground truth normal vectors for pixel $n$ respectively. We will also be considering the proportion of the predictions that fall within 11.25, 22.5, an 30.0 degrees of error.

\begin{equation}
    D_\theta = \text{arccos}(\sum_{n \epsilon N_N}\hat{\textbf{y}}_n\cdot\textbf{y}_n)
\label{angle_distance}
\end{equation}

Finally, for 13-label semantic segmentation, we evaluate the mean intersection over union (mIoU) and the absolute pixel accuracy. In a multi-class setting, the mean intersection over union can be interpreted as taking the mean of the IoU computation in equation \eqref{iou} for every class, where TP, FP, and FN are true positives, false positives, and false negatives respectively.

\begin{equation}
    IoU = \frac{TP}{TP + FP + FN}
\label{iou}
\end{equation}

\begin{table*}[htb!]
    \centering
    \caption{Validation set performance metrics taken across all tasks. The values used are the average performance metrics from the final 10 (out of 200) training epochs. The number of parameters for each model relative to MTAN are also included. Values in boldface indicate the best value in a given column for the multitask learning models.}
    \begin{tabular}{c c c c c c c c c c c c c c} %{\textwidth}
    \toprule
      & & \multicolumn{2}{c}{Segmentation} & & \multicolumn{2}{c}{Depth} & & \multicolumn{5}{c}{Surface Normal} \\ 
     \cline{3-4}
     \cline{6-7}
     \cline{9-13}
      & & & & & & & & \multicolumn{2}{c}{Angle Distance} & \multicolumn{3}{c}{Within t\degree} \\ 
      Model & Rel & \multicolumn{2}{c}{(Higher Better)} & & \multicolumn{2}{c}{(Lower Better)} & & \multicolumn{2}{c}{(Lower Better)} & \multicolumn{3}{c}{(Higher Better)} \\ 
      & Param. & mIoU & Pix Acc & & Abs Err & Rel Err & & Mean & Median & 11.25 & 22.5 & 30\\
     \midrule
     One Task & 1.69 & 27.45 & 54.64 & & 0.6517 & 0.2829 & & 29.85 & 23.67 & 24.83 & 48.34 & 60.17 \\ 
     \midrule
     MTAN  & 1.00 & 28.39 & 55.78 & & 0.6152 & 0.2685 & & 31.57 & 27.50 & 18.68 & 41.40 & 54.51 \\ 
     Split & 1.13 & 15.12 & 40.95 & & 0.7725 & 0.3213 & & 36.81 & 33.37 & 13.70 & 32.89 & 45.07 \\
     PAD-Net & 1.03 & 27.70 & 54.76 & & 0.6393 & 0.2670 & & 33.05 & 28.25 & 19.38 & 40.83 & 52.93 \\
     ATI-Net & 1.03 & \textbf{28.66} & \textbf{56.80} & & \textbf{0.6076} & \textbf{0.2582} & & \textbf{30.95} & \textbf{25.97} & \textbf{21.42} & \textbf{44.23} & \textbf{56.72} \\
    \bottomrule
    \end{tabular}
    \label{tab:training_summary_pretraining}
\end{table*}

\subsection{Baselines}

\s{@Dimitri:  Fill this Section. I think we should not include Split, Wide. What do you think? Then in the table, you would only report the``Split, Dense'' results, call it Split. In this way, the results of ATI-Net will all be in boldface, as is desired.}

The baselines used in our experimentation include three single task networks (one for each task) and three multitask networks. Similar to \cite{liu2019end}, our model can be applied to any feed forward encoder-decoder architecture designed for dense prediction tasks. To make a uniform comparison, all baselines will use the same SegNet backbone.

\begin{itemize}
    \item \textbf{Single-Task, One Task:} This is simply the standard SegNet model for single task learning. In table \ref{tab:training_summary_pretraining}, the results are presented in one row. However, the results for each task were obtained used a separate task-specific model.
    \item \textbf{Multi-Task, Split \cite{liu2019end}:} This is a standard baseline in multitask learning, where all features are shared across tasks until there is a split at the final prediction layer (a.k.a hard parameter sharing). Specifically, this is the same implementation from \cite{liu2019end}.
    \item \textbf{Multi-Task, MTAN \cite{liu2019end}:} This is the SegNet based MTAN taken from the public repository of the authors of \cite{liu2019end}.
    \item \textbf{Multi-Task, PAD-Net  \cite{8578175}:} This is the implementation of PAD-Net presented in \cite{8578175}, but with a SegNet backbone. The features from the shared SegNet encoder are fed to each of the task-specific SegNet decoders to generate the initial predictions. The final predictions are generated after task prediction distillation.
\end{itemize}

\subsection{Implementation Details}
A SegNet \cite{7803544} backbone is used in ATI-Net and all baselines in our experimentation. SegNet is an encoder-decoder architecture for image segmentation, where the encoder consists of a series of convolutions followed by 2$\times$2 max pooling layers. Specifically, the 13 convolutional layers are from VGG-16 \cite{simonyan2014very} with the original fully connected layers removed. The decoder architecture mirrors the encoder in structure while using upsamples and convolutions. The final pixel-wise classification layer from the original SegNet is not included since the backbone is used to provide features to the attention modules and not to make classifications.

MTAN has demonstrated state-of-the-art performance on dense prediction tasks using completely uninitialized network parameters. Therefore, ATI-Net and all comparative models are trained from scratch. To establish a fair testing environment and to adhere to the theme of end-to-end multitask learning, all models were trained for a total of 200 epochs using a batch size of 2. We trained all models using a standard Adam optimizer \cite{kingma2014adam} with an initial learning rate of $10^{-4}$ and a step scheduler that reduces the learning rate by a factor of 0.5 halfway through the training. Additionally, ATI-Net and all multitask models used the DWA \cite{liu2019end} task balancing scheme with a temperature value, $T=2$. As seen in \cite{liu2019end}, DWA determines the weights seen in equation \eqref{naive_loss} in accordance to equation \eqref{weights}, where $\lambda_i(t)$ corresponds to the weight of task $i$ at iteration index $t$. $w_i$ is the relative descending rate \cite{liu2019end}.
\begin{equation}
    \lambda_i(t) = \frac{K \text{exp}(w_i(t-1)/T)}{\sum_k \text{exp}(w_k(t-1)/T)}
\label{weights}
\end{equation}

\begin{equation}
     w_i(t-1) = \frac{\mathcal{L}_i(t-1)}{\mathcal{L}_i(t-2)}
\end{equation}
For the first two iteration indices, $w_i(1) = w_i(2) = 1$. Finally, $K$ is equal to the number of tasks, which is 3 in our case.

All our code was implemented using PyTorch and can be found at the following repository: \url{https://github.com/Armanfard-Lab/ATI-Net} 

\section{Results}
\s{Revise the whole section and also Conclusion section.}
\begin{figure*}[htb!]
\centering
\includegraphics[width=0.8\linewidth]{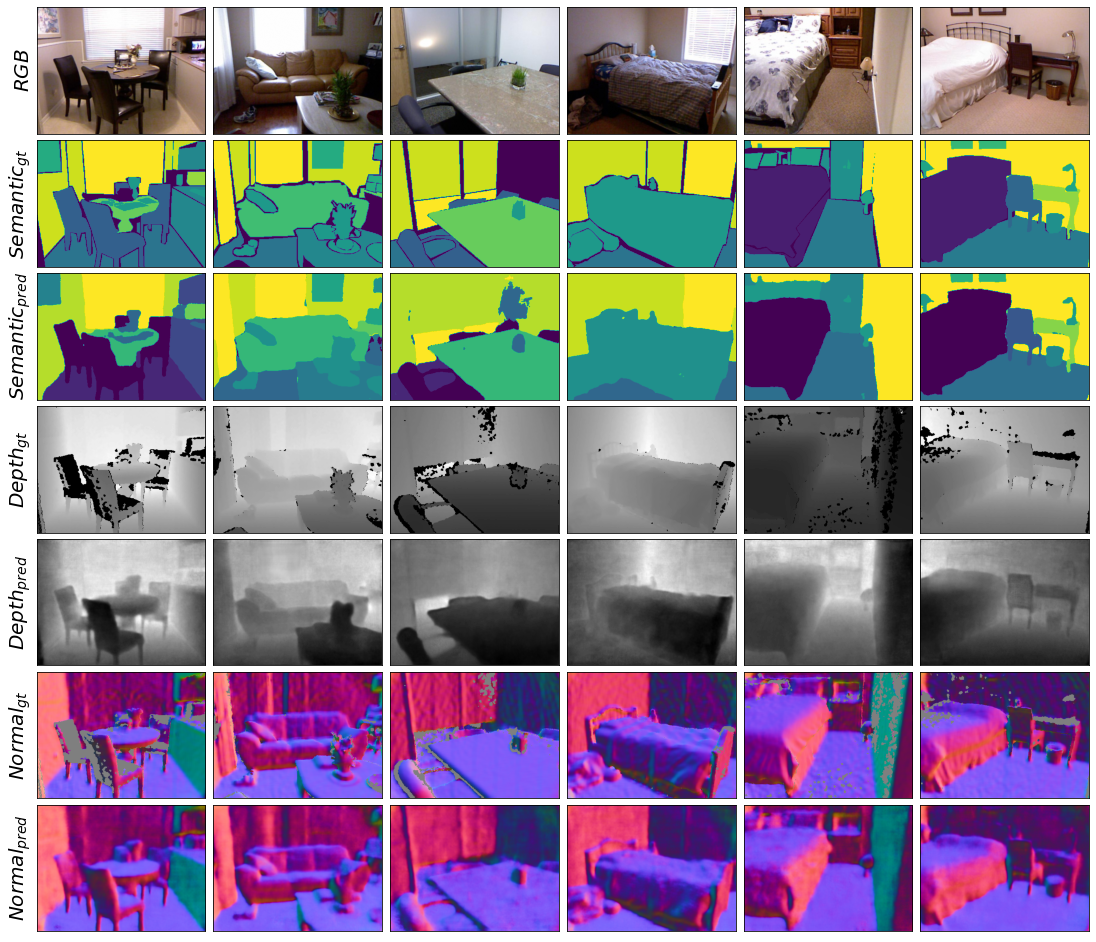}
\caption{Qualitative results obtained using ATI-Net for each task on 6 samples from the NYUv2 dataset. The task labels with the subscript ``gt" and ``pred" correspond to ground truth and predictions from ATI-Net respectively. We can see that despite the missing values in the ground truth data for depth regression and surface normals estimation, the predictions demonstrate the ability to generalize well.}
\label{fig:Qualitative}
\end{figure*}

Table \ref{tab:training_summary_pretraining} summarizes the average validation set results over the last 10 epochs of training for ATI-Net and all baseline models. It also specifies the number of parameters for each model relative to MTAN. MTAN was chosen as the reference point to illustrate how our proposed task interaction system leads to a small change in model parameters. We can see that our proposed ATI-Net outperforms every multitask baseline in every performance metric across each of the tasks. It is also worth noting that our lightweight distillation configuration only introduces approximately 3\% more model parameters compared to MTAN, while obtaining more than 1\% improvement on almost all performance metrics. We can also see that ATI-Net greatly outperforms the single task baselines for semantic segmentation and depth regression. Although, the single task baseline for surface normals estimation outperforms all MTL models, ATI-Net is the closest performing MTL model. The great performance of the single task baseline for surface normals can simply be attributed to the fact it has more model parameters dedicated to a specific task compared to the MTL models. The relative parameter rating for the One Task model is simply the sum of all three individual networks. Consequently, the surface normals model has more than half the number of parameters compared to ATI-Net just to perform a single task.

Figure \ref{fig:Qualitative} contains qualitative results of the predictions generated from ATI-Net for each of the tasks. The RGB images in the first row give a glimpse at the complexity of this indoor scene dataset. Despite its complexity, we can see that ATI-Net generates predictions that resemble the ground truth maps with relatively good clarity and granularity between objects. Even for instances where there are many missing values in the ground truth data, like in the first column, we can see that ATI-Net manages to leverage the domain specific knowledge across tasks to generalize realistic results in the regions with missing data.

\section{Conclusion}
% Not just a stitching of PAD-Net and MTAN (version B is a non-trivial application of TPD)
% Takes benefits from both and improves on drawbacks.
% Improves MTAN performance and takes task interactions into account
% Removes the need for optimizing for one task (i.e extra training) and the need for any additional training at all with version A.
In this study, we proposed ATI-Net, which is an architecture that unites two distinct multi-task learning concepts in a novel way. Specifically, ATI-Net applies task prediction distillation in an attention based multitask learning system. Rather than distilling features at the prediction layer to generate final outputs like other methods, we distill features at an intermediate stage of the network as means to provide better features for the attention masks to extract from the shared backbone. This enhancement to the traditional multitask attention network has led to improvements across all performance metrics for semantic segmentation, depth regression, and surface normals estimation on the challenging NYUv2 dataset. Distilling at an intermediate stage in the network also allows us to distill knowledge between low dimensional features.  Consequently, we introduce very few extra model parameters; unlike other methods that employ expensive distillation algorithms.


\section*{Acknowledgment}
The authors would like to thank the Natural Sciences and Engineering Research Council of Canada and the Department of Electrical and Computer Engineering at McGill University. This work would not have been possible without their financial support.\s{@Dimitri: Also thank Compute Canada and the QC section, if used.} The authors would also like to thank  Calcul Québec and Compute Canada for providing the necessary computational resources to conduct our experiments.




%\bibliographystyle{IEEEtran}


% \end{thebibliography}

\bibliographystyle{IEEEtranN}
\bibliography{References}

% Generated by IEEEtranN.bst, version: 1.14 (2015/08/26)
\begin{thebibliography}{22}
\providecommand{\natexlab}[1]{#1}
\providecommand{\url}[1]{#1}
\csname url@samestyle\endcsname
\providecommand{\newblock}{\relax}
\providecommand{\bibinfo}[2]{#2}
\providecommand{\BIBentrySTDinterwordspacing}{\spaceskip=0pt\relax}
\providecommand{\BIBentryALTinterwordstretchfactor}{4}
\providecommand{\BIBentryALTinterwordspacing}{\spaceskip=\fontdimen2\font plus
\BIBentryALTinterwordstretchfactor\fontdimen3\font minus
  \fontdimen4\font\relax}
\providecommand{\BIBforeignlanguage}[2]{{%
\expandafter\ifx\csname l@#1\endcsname\relax
\typeout{** WARNING: IEEEtranN.bst: No hyphenation pattern has been}%
\typeout{** loaded for the language `#1'. Using the pattern for}%
\typeout{** the default language instead.}%
\else
\language=\csname l@#1\endcsname
\fi
#2}}
\providecommand{\BIBdecl}{\relax}
\BIBdecl

\bibitem[Caruana(1997)]{caruana1997multitask}
R.~Caruana, ``Multitask learning,'' \emph{Machine learning}, vol.~28, no.~1,
  pp. 41--75, 1997.

\bibitem[Kendall et~al.(2018)Kendall, Gal, and Cipolla]{kendall2018multi}
A.~Kendall, Y.~Gal, and R.~Cipolla, ``Multi-task learning using uncertainty to
  weigh losses for scene geometry and semantics,'' in \emph{Proceedings of the
  IEEE conference on computer vision and pattern recognition}, 2018, pp.
  7482--7491.

\bibitem[Liu et~al.(2019)Liu, Johns, and Davison]{liu2019end}
S.~Liu, E.~Johns, and A.~J. Davison, ``End-to-end multi-task learning with
  attention,'' in \emph{Proceedings of the IEEE/CVF Conference on Computer
  Vision and Pattern Recognition}, 2019, pp. 1871--1880.

\bibitem[Standley et~al.(2020)Standley, Zamir, Chen, Guibas, Malik, and
  Savarese]{standley2020tasks}
T.~Standley, A.~Zamir, D.~Chen, L.~Guibas, J.~Malik, and S.~Savarese, ``Which
  tasks should be learned together in multi-task learning?'' in
  \emph{International Conference on Machine Learning}.\hskip 1em plus 0.5em
  minus 0.4em\relax PMLR, 2020, pp. 9120--9132.

\bibitem[Xu et~al.(2018)Xu, Ouyang, Wang, and Sebe]{8578175}
D.~Xu, W.~Ouyang, X.~Wang, and N.~Sebe, ``Pad-net: Multi-tasks guided
  prediction-and-distillation network for simultaneous depth estimation and
  scene parsing,'' in \emph{2018 IEEE/CVF Conference on Computer Vision and
  Pattern Recognition}, 2018, pp. 675--684.

\bibitem[{Vandenhende} et~al.(2020){Vandenhende}, {Georgoulis}, and
  {Gool}]{vandenhende2020mti}
S.~{Vandenhende}, S.~{Georgoulis}, and L.~V. {Gool}, ``Mti-net: Multi-scale
  task interaction networks for multi-task learning,'' in \emph{European
  Conference on Computer Vision}, 2020, pp. 527--543.

\bibitem[Eigen and Fergus(2015)]{eigen2015predicting}
D.~Eigen and R.~Fergus, ``Predicting depth, surface normals and semantic labels
  with a common multi-scale convolutional architecture,'' in \emph{Proceedings
  of the IEEE international conference on computer vision}, 2015, pp.
  2650--2658.

\bibitem[Kokkinos(2017)]{8100062}
I.~Kokkinos, ``Ubernet: Training a universal convolutional neural network for
  low-, mid-, and high-level vision using diverse datasets and limited
  memory,'' in \emph{2017 IEEE Conference on Computer Vision and Pattern
  Recognition (CVPR)}, 2017, pp. 5454--5463.

\bibitem[Sermanet et~al.(2013)Sermanet, Eigen, Zhang, Mathieu, Fergus, and
  LeCun]{sermanet2013overfeat}
P.~Sermanet, D.~Eigen, X.~Zhang, M.~Mathieu, R.~Fergus, and Y.~LeCun,
  ``Overfeat: Integrated recognition, localization and detection using
  convolutional networks,'' \emph{arXiv preprint arXiv:1312.6229}, 2013.

\bibitem[Kendall and Gal(2017)]{kendall2017uncertainties}
A.~Kendall and Y.~Gal, ``What uncertainties do we need in bayesian deep
  learning for computer vision?'' \emph{arXiv preprint arXiv:1703.04977}, 2017.

\bibitem[Chen et~al.(2018)Chen, Badrinarayanan, Lee, and
  Rabinovich]{chen2018gradnorm}
Z.~Chen, V.~Badrinarayanan, C.-Y. Lee, and A.~Rabinovich, ``Gradnorm: Gradient
  normalization for adaptive loss balancing in deep multitask networks,'' in
  \emph{International Conference on Machine Learning}.\hskip 1em plus 0.5em
  minus 0.4em\relax PMLR, 2018, pp. 794--803.

\bibitem[Guo et~al.(2018)Guo, Haque, Huang, Yeung, and Fei-Fei]{guo2018dynamic}
M.~Guo, A.~Haque, D.-A. Huang, S.~Yeung, and L.~Fei-Fei, ``Dynamic task
  prioritization for multitask learning,'' in \emph{Proceedings of the European
  Conference on Computer Vision (ECCV)}, 2018, pp. 270--287.

\bibitem[Vandenhende et~al.(2021)Vandenhende, Georgoulis, Van~Gansbeke,
  Proesmans, Dai, and Van~Gool]{9336293}
S.~Vandenhende, S.~Georgoulis, W.~Van~Gansbeke, M.~Proesmans, D.~Dai, and
  L.~Van~Gool, ``Multi-task learning for dense prediction tasks: A survey,''
  \emph{IEEE Transactions on Pattern Analysis and Machine Intelligence}, pp.
  1--1, 2021.

\bibitem[Ruder(2017)]{ruder2017overview}
S.~Ruder, ``An overview of multi-task learning in deep neural networks,''
  \emph{arXiv preprint arXiv:1706.05098}, 2017.

\bibitem[Zhang et~al.(2018)Zhang, Cui, Xu, Jie, Li, and Yang]{zhang2018joint}
Z.~Zhang, Z.~Cui, C.~Xu, Z.~Jie, X.~Li, and J.~Yang, ``Joint task-recursive
  learning for semantic segmentation and depth estimation,'' in
  \emph{Proceedings of the European Conference on Computer Vision (ECCV)},
  2018, pp. 235--251.

\bibitem[Zhang et~al.(2019)Zhang, Cui, Xu, Yan, Sebe, and
  Yang]{zhang2019pattern}
Z.~Zhang, Z.~Cui, C.~Xu, Y.~Yan, N.~Sebe, and J.~Yang, ``Pattern-affinitive
  propagation across depth, surface normal and semantic segmentation,'' in
  \emph{Proceedings of the IEEE/CVF Conference on Computer Vision and Pattern
  Recognition}, 2019, pp. 4106--4115.

\bibitem[Misra et~al.(2016)Misra, Shrivastava, Gupta, and
  Hebert]{misra2016cross}
I.~Misra, A.~Shrivastava, A.~Gupta, and M.~Hebert, ``Cross-stitch networks for
  multi-task learning,'' in \emph{Proceedings of the IEEE conference on
  computer vision and pattern recognition}, 2016, pp. 3994--4003.

\bibitem[Gao et~al.(2019)Gao, Ma, Zhao, Liu, and Yuille]{gao2019nddr}
Y.~Gao, J.~Ma, M.~Zhao, W.~Liu, and A.~L. Yuille, ``Nddr-cnn: Layerwise feature
  fusing in multi-task cnns by neural discriminative dimensionality
  reduction,'' in \emph{Proceedings of the IEEE/CVF Conference on Computer
  Vision and Pattern Recognition}, 2019, pp. 3205--3214.

\bibitem[Nathan~Silberman and Fergus(2012)]{Silberman:ECCV12}
P.~K. Nathan~Silberman, Derek~Hoiem and R.~Fergus, ``Indoor segmentation and
  support inference from rgbd images,'' in \emph{ECCV}, 2012.

\bibitem[Badrinarayanan et~al.(2017)Badrinarayanan, Kendall, and
  Cipolla]{7803544}
V.~Badrinarayanan, A.~Kendall, and R.~Cipolla, ``Segnet: A deep convolutional
  encoder-decoder architecture for image segmentation,'' \emph{IEEE
  Transactions on Pattern Analysis and Machine Intelligence}, vol.~39, no.~12,
  pp. 2481--2495, 2017.

\bibitem[Simonyan and Zisserman(2014)]{simonyan2014very}
K.~Simonyan and A.~Zisserman, ``Very deep convolutional networks for
  large-scale image recognition,'' \emph{arXiv preprint arXiv:1409.1556}, 2014.

\bibitem[Kingma and Ba(2014)]{kingma2014adam}
D.~P. Kingma and J.~Ba, ``Adam: A method for stochastic optimization,''
  \emph{arXiv preprint arXiv:1412.6980}, 2014.

\end{thebibliography}

\end{document}